\documentclass[final]{cvpr}

\usepackage{times}
\usepackage{epsfig}
\usepackage{graphicx}
\usepackage{amsmath}
\usepackage{amssymb}
\usepackage[table]{xcolor}
\usepackage{multirow}
\usepackage{algorithmic}
\usepackage{subcaption}

\newsavebox{\ieeealgbox}

\usepackage[pagebackref=true,breaklinks=true,colorlinks,bookmarks=false]{hyperref}

\def\cvprPaperID{12} 
\def\confYear{CVPR 2021}
\begin{document}

\title{\vspace{-0.5cm} Evaluating the Robustness of Bayesian Neural Networks Against Different Types of Attacks}

\author{Yutian Pang, Sheng Cheng, Jueming Hu, Yongming Liu \\ 
Arizona State University, Tempe, AZ \\ 
{\tt\small \{yutian.pang, scheng53, jueming.hu, yongming.liu\}@asu.edu}\\[-0.5ex]
}

\maketitle

\begin{abstract}
   To evaluate the robustness gain of Bayesian neural networks on image classification tasks, we perform input perturbations, and adversarial attacks to the state-of-the-art Bayesian neural networks, with a benchmark CNN model as reference. The attacks are selected to simulate signal interference and cyberattacks towards CNN-based machine learning systems. The result shows that a Bayesian neural network achieves significantly higher robustness against adversarial attacks generated against a deterministic neural network model, without adversarial training. The Bayesian posterior can act as the safety precursor of ongoing malicious activities. Furthermore, we show that the stochastic classifier after the deterministic CNN extractor has sufficient robustness enhancement rather than a stochastic feature extractor before the stochastic classifier. This advises on utilizing stochastic layers in building decision-making pipelines within a safety-critical domain. 
\end{abstract}

\section{Introduction \label{sec: introduction}}

Deep Neural Networks (DNNs) have been integrated into various safety-critical engineering applications (\eg Unmanned Aerial Vehicle (UAV), Autonomous System (AS), Surveillance System (SS)). The prediction made by these algorithms needs to be reliable with sufficient robustness. A failed DNN can lead to potentially fatal collisions, especially for the solely camera-based autonomous systems. Several such real-world accidents have happened including ones that resulted in a fatality \cite{tian2018deeptest}, where the image of the white-colored truck was classified as the cloud. On the other hand, it's widely known that the predicted labels of neural networks are vulnerable to adversarial samples \cite{agarwal2019improving, dong2017towards, goodfellow2014explaining}. The research on adversarial machine learning has focused on developing an enormous number of adversarial attack and defense methods \cite{carlini2017adversarial, carlini2017towards, zhang2019limitations}.  Most of the adversarial attack/defense methods are developed towards the classical convolutional neural network (CNN) on image classification tasks. Typically, the defense requires adversarial training with adversarial samples that are used to perform adversarial attacks. In this way, the model can act more robust against these types of attacks. However, it's worth pointing out that the development of new attack methods never ends.

Bayesian NNs, with distributions over their weights, are gaining attention for their uncertainty quantification ability and high robustness from Bayesian regularization, while retaining the advantages of deterministic NNs \cite{cardelli2019statistical}. The robustness gain of BNNs is not rigorous studied in the literature yet lacking quantified comparative experiments on a real-world dataset. In particular, we compare various types of Bayesian inference methods to NNs including Bayes By Backprop (BBB) \cite{blundell2015weight} with (local) reparameterization \cite{kingma2015variational, molchanov2017variational}, Variational Inference (VI) \cite{kingma2013auto}, and Flipout approximation \cite{wen2018flipout}. BNNs are evaluated against several types of input perturbations, white-box adversarial attacks, and black-box adversarial attacks without adversarial training. These attacks simulate the possible attacks toward a deployed NN system in the real world, intentionally or unintentionally. The adversarial samples are generated with the $L_p$ threat models. In this paper, we adopt 6 input perturbation methods, 5 white-box adversarial attacks, 3 black-box attacks towards two open-source datasets (German Traffic Sign Recognition (GTSRB) \cite{Houben-IJCNN-2013} \& Planes in Satellite Imagery (PlanesNet) \cite{planes}), both of which were in the safety-critical domains (AS \& SS).

We have several exciting findings by analyzing the experiment results quantitatively. 
Firstly, we notice that BNN has limited robustness benefits against various input perturbations since the classical CNN has also demonstrated denoising capabilities
Secondly, the Bayesian neural network shows significant robustness in the experiments in terms of classification accuracy, especially against constrained adversarial attacks \cite{dong2020benchmarking}.
Thirdly, 
we realize that both models will fail when dealing with unconstrained adversarial attacks. In this case, the attacks are obviously distinguishable by human visions.
Furthermore, the stochasticity on the classifier can achieve comparative performance by putting weight uncertainties on both the convolutional extractor and the classifier, with comparative computation time consumption. 
These findings give advice on building robust stochastic image-based classifiers in real-world machine learning system applications. More discussions are presented in Sec~\ref{sec: conclusion}. 


\begin{table*}[ht!]
\center
\caption{Performance Against Input Perturbations. Report Test Accuracy in \%.}
\label{table1}

\scalebox{0.85}{
\noindent\begin{tabular}{|p{0.05\linewidth}|p{0.07\linewidth}|p{0.1\linewidth}|p{0.1\linewidth}|p{0.1\linewidth}|p{0.1\linewidth}|p{0.1\linewidth}|p{0.1\linewidth}|p{0.1\linewidth}|}
\hline
\rowcolor{lightgray}\multicolumn{9}{|l|}{\textbf{Dataset I: GTSRB(German Traffic Sign Recognition Benchmark)}} \\
\hline
        \multicolumn{2}{|l|}{Methods} & Clean & Gaussian & S\&P & Poisson & RE & RE Colorful & Speckle\\ 
        \hline\hline
        CNN & Baseline & $96.28\pm 0.69$ &$96.19\pm0.72$ &$76.91\pm4.37$ &$96.29\pm0.69$ &$88.76\pm0.79$ &$72.50\pm4.52$ &$15.18\pm2.33$ \\
        \hline\hline
        \multirow{ 2}{*}{F-BNN} &Flipout &$97.17\pm0.22$& $97.19\pm0.17$ & $84.06\pm1.00$& $97.18\pm0.17 $& $90.45\pm 0.33$& $82.12\pm0.55$& $34.12\pm3.27$\\
        \cline{2-9}
        & BBB &$97.25\pm 0.16$ &$97.27\pm0.16 $ &$80.81\pm1.61$ &$97.26\pm0.21 $ &$89.81\pm 0.92$ &$79.69\pm1.99$ &$26.89\pm2.81$ \\
        \hline\hline
        \multirow{4}{*}{BNN} &Flipout &$96.85\pm 0.51$ &$96.87\pm0.56$ &$75.51\pm4.49$ &$96.90\pm0.55$ &$89.93\pm0.85 $ &$72.96\pm1.60$ &$10.46\pm4.52 $ \\
        \cline{2-9}
        &BBB &$96.93\pm0.32$ &$96.94\pm0.35 $ &$79.42\pm1.93 $ &$96.93\pm0.34$ &$88.45\pm1.32 $ &$73.65\pm2.71 $ &$17.48\pm2.97$ \\
        \cline{2-9}
        &LRT &$96.85\pm0.16$ &$96.83\pm0.20$ &$75.60\pm1.13$ &$96.83\pm0.23$ &$88.64\pm 1.22$ &$71.88\pm4.22$ &$10.88\pm2.50$ \\
        \cline{2-9}
        &VI &$95.37\pm 0.41$ &$95.30\pm0.44$ &$73.33\pm3.65 $ &$95.35\pm0.40 $ &$85.83\pm1.84$ &$68.91\pm0.99 $ &$10.05\pm2.42$ \\
        \hline 
\end{tabular}
}
\end{table*}
\section{Background \label{sec: background}}

\subsection{Bayesian neural networks \label{subsec: bnn}}
The formulation of Bayesian NN relies on Bayesian probabilistic modeling with i.i.d. distributions over network parameters. The Bayesian approach gives a space of parameters $\omega$ as a distribution $p(\omega)$ called the \textit{prior}, and a likelihood distribution $p(Y|X, \omega)$, which is a probabilistic model of the model outputs given $X$ and $\omega$. The posterior is proportional to the likelihood and the prior and the prediction is simply $\mathbb{E}_{p(\omega|X, Y)}[p(Y^*|X^*, \omega)]$. $X^*$ is the test input and $Y^*$ is the prediction. However, the inference of the posterior and the prediction are both intractable \cite{blundell2015weight}.

Variational Inference (VI) \cite{kingma2013auto}, as an approximate probabilistic inference method, is used to resolve this. The objective is to minimize the distance between the approximate variational distribution $q_\theta(\omega)$ for the posterior $p(\omega|X,Y)$. 
The objective is further approximated as the negative Evidence Lower BOund (ELBO). 
In practice, ELBO is approximated by $\sum^{n}_{i=1} [q_{\omega(i)}(\omega) - \log p(\omega^{(i)}) - \log p(Y|X, \omega^{(i)})]$, where $\omega^{(i)}$ is the $i^{th}$ Monte Carlo sample from $q_\theta(\omega)$. $\omega$ is reparameterized into $(\mu, \sigma)$ for backpropagation \cite{kingma2015variational, molchanov2017variational}. Sampling the network parameters stochastically during training is referred as \textit{weight perturbations}. The recent advancements of weight perturbation method, Flipout, decorrelate the gradients within each batch of the data \cite{wen2018flipout}, while boosting the inference process of BNN.

\subsection{Adversarial Attacks \label{subsec: adversarial}}
Adversarial attacks are defined based upon the concept of threat model \cite{carlini2019evaluating}. 
Denote $f(\cdot)$ as a classification model, with original input $x$ and adversarial samples $x^{adv}$, and $y$ denotes the ground-truth label. 
The adversarial attack is to attack the model $f$ by adding small perturbation on the original inputs.
This perturbation measured by the $L_p$ norm is limited by the perturbation budget $\varepsilon$, that is $\|x - x^{adv}\|_p < \varepsilon$. 
Particularly, we use $L_\infty$ in this paper which implies that the perturbation to each pixel in $x$ can't be larger than $\varepsilon$. The generation of adversarial samples is formulated into two optimization problems depend on if $\varepsilon$ presented. Eq.~\eqref{op1} is to generate an untargeted adversarial example by maximizing the cross-entropy loss function $\mathcal{L}$. The second strategy is to find the minimum perturbation as Eq.~\eqref{op2}. $x'$ is one proposed adversarial sample by the generation algorithm.

\begin{equation}
\label{op1}
    x^{adv} \leftarrow \underset{\|x - x'\|_p < \varepsilon}{\mathrm{argmax}} \mathcal{L}(x', y)
\end{equation}
\begin{equation}
\label{op2}
    x^{adv} \leftarrow \underset{x'}{\mathrm{argmin}} \|x - x'\|_p
\end{equation}

The concept of white-box attacks and black-box attacks build upon the level of adversary's knowledge. White-box attacks typically acquire the full knowledge of the model, including model architectures, parameters, loss formula. White-box attack methods generate perturbations based on NN gradients given the detailed knowledge of the model. Under Eq.~\eqref{op1}, the Fast Gradient Sign Method (FGSM) \cite{goodfellow2014explaining} generates $x^{adv}$ by an one-step update. Basic Iterative Method (BIM) \cite{kurakin2016adversarial} is an iterative version of FGSM with a multi-step update. Projected gradient descent method (PGD) \cite{madry2017towards} has a similar first-order setup but with random initials. DeepF \cite{moosavi2016deepfool} and Carlini \& Wagner's method (C\&W) \cite{carlini2017towards} have been used to solve Eq.~\eqref{op2}. 

Black-box attacks have limit/partial knowledge of the target model. Depending on the portion of the knowledge to the model, it can be further categorized into transfer-based, score-based, and decision-based black-box attacks. Transfer-based attack uses \textit{distillation as a defense} strategy by training a substitute model with the knowledge of the training data. Momentum Iterative Method (MIM) \cite{dong2018boosting} gives guidance on update direction as an extension of BIM. Score-based black-box attacks only acquire the output probabilities. It estimates the gradients by gradient-free methods with limited queries. An example of a score-based attack is SPSA \cite{uesato2018adversarial}. Decision-based black-box attacks solely acquire the hard-label predictions. The Square Attack \cite{andriushchenko2020square} is an example based on a random search on the decision boundary.

\subsection{Input perturbations \label{subsec: perturbations}}
Input perturbations to $x$, as a similar concept to data augmentation, are also examined in this paper as they also exist in real-world cases. We adopt 6 types of input perturbation methods to simulate various user cases. Firstly, we use the recent advancement Random Erasing (RE) \cite{zhong2020random}. RE randomly masks a rectangular region with black color in $x$ with several masking parameters to determine the size of the region. Furthermore, we generate RE with random color as masking of the inputs. A typical case is stickers on the stop sign and fails a self-driving car. We also adopt the salt-and-pepper noise \cite{azzeh2018salt}, and speckle noise to simulate signal interference. This includes the electromagnetic interference (EMI) in unshielded twisted pairs (UTP) in Ethernet or adjacent-channel interference in frequency modulation (FM) systems. We use Gaussian/Possion blur on $x$ to simulate the system with low data transmission speed, and/or bandwidth issues.

\section{Experiments \label{sec: experiments}}

\subsection{Evaluated Datasets \label{data}}
We use the GTSRB \cite{Houben-IJCNN-2013} and PlanesNet \cite{planes} in this paper. The GTSRB images are classified into 43 classes where the training set contains $39,209$ images and $12,630$ in test set. The PlansNet has two class labels indicate \textit{plane} or \textit{no-plane} given a input image. We use $10\%$ of the data for testing. The attack methods discussed in Sec. \ref{subsec: adversarial} and Sec. \ref{subsec: perturbations} are used in the experiment. The visualization of test images are shown in Appendix. 


\subsection{Evaluation Procedures \label{procedures}}

Firstly, we train the baseline CNN model, the F-BNN model, and the BNN model. We build all the models following the VGG-16 architecture, but with stochastic layers in the Bayesian formulation. F-BNN here refers to fully-Bayesian neural network where both feature extractor and classifier are stochastic, while only stochastic classifier presented in BNN model. Then, we generate different types of adversarial samples of the test data w.r.t. the baseline CNN model and test each of the trained model to get the classification accuracy against adversarial attacks. Repeating this procedure for 5 times and averaging the results to get the mean prediction accuracy and variance. The evaluation procedure for input perturbations are similar to this procedure. We report the quantitative results in Table.~\ref{table1}, Table.~\ref{table2}.



\begin{table*}[ht!]
\center

\caption{Performance Against Adversarial Attacks with different $\varepsilon$ and dataset. Report Test Accuracy in $\%$.}
\label{table2}

\scalebox{0.8}{
\noindent\begin{tabular}{|p{0.06\linewidth}|p{0.07\linewidth}|p{0.1\linewidth}|p{0.1\linewidth}|p{0.1\linewidth}|p{0.1\linewidth}|p{0.1\linewidth}|p{0.1\linewidth}|p{0.1\linewidth}|p{0.1\linewidth}|}
\hline
\rowcolor{lightgray}\multicolumn{10}{|l|}{\textbf{Dataset I: GTSRB(German Traffic Sign Recognition Benchmark)}} \\
\hline
        \multicolumn{2}{|l|}{$L_{\infty}(\epsilon=.10$)}  & \multicolumn{5}{l}{White-box Attacks} & \multicolumn{3}{|l|}{Black-box Attacks}\\\hline
        \multicolumn{2}{|l|}{Methods/Distance}& PGD/$0.056$ & FGSM/$0.071$ & BIM/$0.049$ & C\&W/$0.001$ & DeepF/$0.283$ & SPSA/$0.290$ & MIM/$0.092$ &Square/$0.088$\\ 
        \hline\hline
        CNN & Baseline & $2.43\pm0.50 $ &$28.79\pm1.69 $ &$28.35\pm 1.65$ &$83.80\pm6.19$ & $3.97\pm0.64$ & $3.27\pm0.12$ &$2.26\pm0.41$ & $15.26\pm3.05$\\
        \hline\hline
        \multirow{2}{*}{F-BNN} &Flipout & $77.86\pm2.57 $ &$58.86\pm1.68 $ &$68.08\pm2.53 $ &$95.59\pm0.27$ & $20.58\pm1.73$ & $3.75\pm0.04$ &$61.59\pm3.00$ & $85.65\pm4.36$\\
        \cline{2-10}
        & BBB & $78.78\pm3.27 $ &$59.96\pm2.01 $ &$69.32\pm2.88 $ &$95.76\pm0.23$ & $20.57\pm1.90$ & $3.78\pm0.04$ &$62.86\pm3.47$ & $80.19\pm3.37$\\
        \hline\hline
        \multirow{4}{*}{BNN} &Flipout & $75.55\pm2.93 $ &$55.71\pm2.39 $ &$65.77\pm2.42 $ &$94.99\pm0.61$ & $18.57\pm1.88$ & $3.78\pm0.06$ &$57.30\pm2.84$ & $70.54\pm5.01$\\
        \cline{2-10}
        &BBB & $76.47\pm1.68 $ &$57.13\pm1.61 $ &$66.69\pm1.58 $ &$95.44\pm0.53$ & $19.47\pm2.65$ & $3.76\pm0.05$ &$58.49\pm2.20$ & $71.65\pm7.76$\\
        \cline{2-10}
        &LRT & $76.78\pm2.04 $ &$57.10\pm1.86 $ &$67.25\pm1.89 $ &$95.55\pm0.35$ & $19.33\pm1.42$ & $3.77\pm0.05$ &$58.90\pm2.80$ & $76.53\pm5.79$\\
        \cline{2-10}
        &VI & $73.14\pm1.89 $ &$53.37\pm1.16 $ &$63.51\pm1.13 $ &$93.85\pm0.59$ & $17.79\pm2.28$ & $3.75\pm0.08$ &$53.97\pm2.18$ & $67.86\pm8.42$\\
        \hline 
\end{tabular}
}

\scalebox{0.8}{
\noindent\begin{tabular}{|p{0.06\linewidth}|p{0.07\linewidth}|p{0.1\linewidth}|p{0.1\linewidth}|p{0.1\linewidth}|p{0.1\linewidth}|p{0.1\linewidth}|p{0.1\linewidth}|p{0.1\linewidth}|p{0.1\linewidth}|}
\hline
\rowcolor{lightgray}\multicolumn{10}{|l|}{\textbf{Dataset I: GTSRB(German Traffic Sign Recognition Benchmark)}} \\
\hline
        \multicolumn{2}{|l|}{$L_{\infty}(\epsilon=.15$)}  & \multicolumn{5}{l}{White-box Attacks} & \multicolumn{3}{|l|}{Black-box Attacks}\\\hline
        \multicolumn{2}{|l|}{Methods/Distance}& PGD/$0.061$ & FGSM/$0.104$ & BIM/$0.060$ & C\&W/$0.002$ & DeepF/$0.285$ & SPSA/$0.293$ & MIM/$0.133$ &Square/$0.132$\\ 
        \hline\hline
        CNN & Baseline & $2.41\pm 0.51$ &$28.29\pm 1.77$ &$28.32\pm1.66 $ &$79.77\pm2.49$ & $3.99\pm 0.64$ & $3.20\pm0.09$ &$2.21\pm0.39$ &$5.24\pm1.66$ \\
        \hline\hline
        \multirow{2}{*}{F-BNN} &Flipout & $73.28\pm2.97 $& $42.79\pm3.68$& $56.20\pm 4.42$& $94.94\pm 0.14$&$20.96\pm0.84$ & $3.80\pm0.02$ &$42.86\pm4.41$ &$73.10\pm5.62$ \\
        \cline{2-10}
        & BBB &$70.00\pm2.37$ &$41.91\pm1.84 $ & $53.87\pm2.13$&$94.97\pm0.14$ &$19.92\pm1.78 $& $3.87\pm0.04$ &$39.76\pm1.79$ &$74.51\pm5.97$  \\
        \hline\hline
        \multirow{4}{*}{BNN} &Flipout & $69.96\pm3.26 $& $40.88\pm0.82 $& $54.65\pm1.25 $&$94.65\pm0.49 $ &$20.01\pm0.27 $ & $3.75\pm0.04$ &$37.58\pm2.41$ &$62.82\pm4.08$ \\
        \cline{2-10}
        &BBB &$70.21\pm1.75$ &$40.92\pm1.41 $ &$54.23\pm1.53 $ & $94.72\pm0.38 $&$19.89\pm 1.13$& $3.83\pm0.04$ &$38.54\pm1.89$ &$58.18\pm9.47$  \\
        \cline{2-10}
        &LRT & $69.39\pm2.20$&$40.48\pm1.34$ &$53.54\pm1.44$ &$94.76\pm0.09$ &$19.11\pm2.36$& $3.78\pm0.07$ &$36.34\pm1.81$ &$60.43\pm11.1$  \\
        \cline{2-10}
        &VI &$67.35\pm 3.84$ &$39.89\pm 1.89$ &$53.28\pm 2.45$ &$93.23\pm0.42 $ & $18.07\pm 1.75$& $3.79\pm0.05$ &$35.12\pm2.43$ &$53.86\pm8.77$ \\
        \hline 
\end{tabular}
}
\scalebox{0.8}{
\begin{tabular}{|p{0.06\linewidth}|p{0.07\linewidth}|p{0.1\linewidth}|p{0.1\linewidth}|p{0.1\linewidth}|p{0.1\linewidth}|p{0.1\linewidth}|p{0.1\linewidth}|p{0.1\linewidth}|p{0.1\linewidth}|}
\hline
\rowcolor{lightgray}\multicolumn{10}{|l|}{\textbf{Dataset II: PlanesNet(Detect Aircraft in Planet Satellite Image Chips)}} \\
\hline
        \multicolumn{2}{|l|}{$L_{\infty}(\epsilon=.10$)}  & \multicolumn{5}{l}{White-box Attacks} & \multicolumn{3}{|l|}{Black-box Attacks}\\\hline
        \multicolumn{2}{|l|}{Methods/Distance} & PGD/$0.068$ & FGSM/$0.086$ & BIM/$0.061$ & C\&W/$0.005$ & DeepF/$0.349$ & SPSA/$0.212$ & MIM/$0.094$ &Square/$0.080$\\ 
        \hline\hline
        CNN & Baseline & $1.81\pm 0.37$ &$45.44\pm 2.40$ &$13.50\pm2.28 $ &$68.21\pm2.13$ & $43.77\pm4.83$ & $49.65\pm0.91$ &$1.83\pm0.36$ & $15.46\pm1.70$\\
        \hline\hline
        \multirow{2}{*}{F-BNN} &Flipout & $24.51\pm3.70 $ &$57.69\pm2.37 $ &$36.78\pm4.35 $ &$91.13\pm0.41$ & $54.90\pm2.84$ & $60.88\pm0.79$ &$23.14\pm2.71$ & $75.36\pm6.81$\\
        \cline{2-10}
        & BBB & $28.43\pm 5.81$ &$58.51\pm2.65 $ &$40.96\pm6.94 $ &$89.69\pm0.82$ & $48.54\pm8.92$ & $66.54\pm1.06$ &$26.28\pm5.04$ & $76.20\pm7.44$\\
        \hline\hline
        \multirow{4}{*}{BNN} &Flipout & $23.54\pm 2.67$ &$62.77\pm1.30 $ &$40.31\pm 3.33$ &$91.33\pm0.76$ & $68.30\pm1.33$ & $65.58\pm2.68$ &$24.07\pm2.25$ & $75.89\pm2.65$\\
        \cline{2-10}
        &BBB & $22.58\pm4.24 $ &$59.74\pm4.23 $ &$38.69\pm8.71 $ &$91.59\pm1.35$ & $61.85\pm5.06$ & $62.94\pm2.94$ &$22.72\pm4.20$ & $74.98\pm4.11$\\
        \cline{2-10}
        &LRT & $25.22\pm1.80 $ &$62.59\pm1.09 $ &$42.23\pm3.93$ &$91.47\pm1.42$ & $69.33\pm2.28$ & $65.57\pm2.16$ &$25.83\pm2.42$ & $77.79\pm1.59$\\
        \cline{2-10}
        &VI & $20.72\pm5.56 $ &$55.79\pm3.38 $ &$33.92\pm 6.17$ &$91.19\pm1.70$ & $65.89\pm4.76$ & $60.30\pm2.49$ &$19.91\pm5.57$ & $72.56\pm4.02$\\
        \hline 
\end{tabular}
}

\end{table*}
\section{Results \label{sec: results}}

Table.~\ref{table1} lists the quantitative results of each model setups against input perturbations. Results with a clean test input show the CNN baseline is well-trained. Variational Bayes performs slightly worse. We analysis input perturbations by groups, a) The Gaussian/Possion noisy blur to the inputs won't affect the model performance. The reason is neural network has been proved to have denoising capabilities, especially for parameterized distributional noise. b) The F-BNN with the RE/Colorful RE input perturbations has better performance among all cases. c) The S\&P/Speckle signal interference cases has the best attack performance.

In Table.~\ref{table2}, we report the results of different models against adversarial attacks on two datasets. For GTSRB dataset, We generate the untargeted adversarial samples using the $L_\infty$ threat model with $\varepsilon=0.10$ and $\varepsilon=0.15$ on both of these two datasets. We perform attacks as previously discussed in Sec.~\ref{subsec: adversarial}. 
We also evaluate the $L_\infty$ norm between adversarial samples and the original inputs for each run. We report the mean $L_\infty$ distance between $x$ and $x^{adv}$ for different attack methods in the table (\eg PGD/0.068).


\raggedbottom

\begin{figure}[ht!]
\centering
\includegraphics[width=1\linewidth]{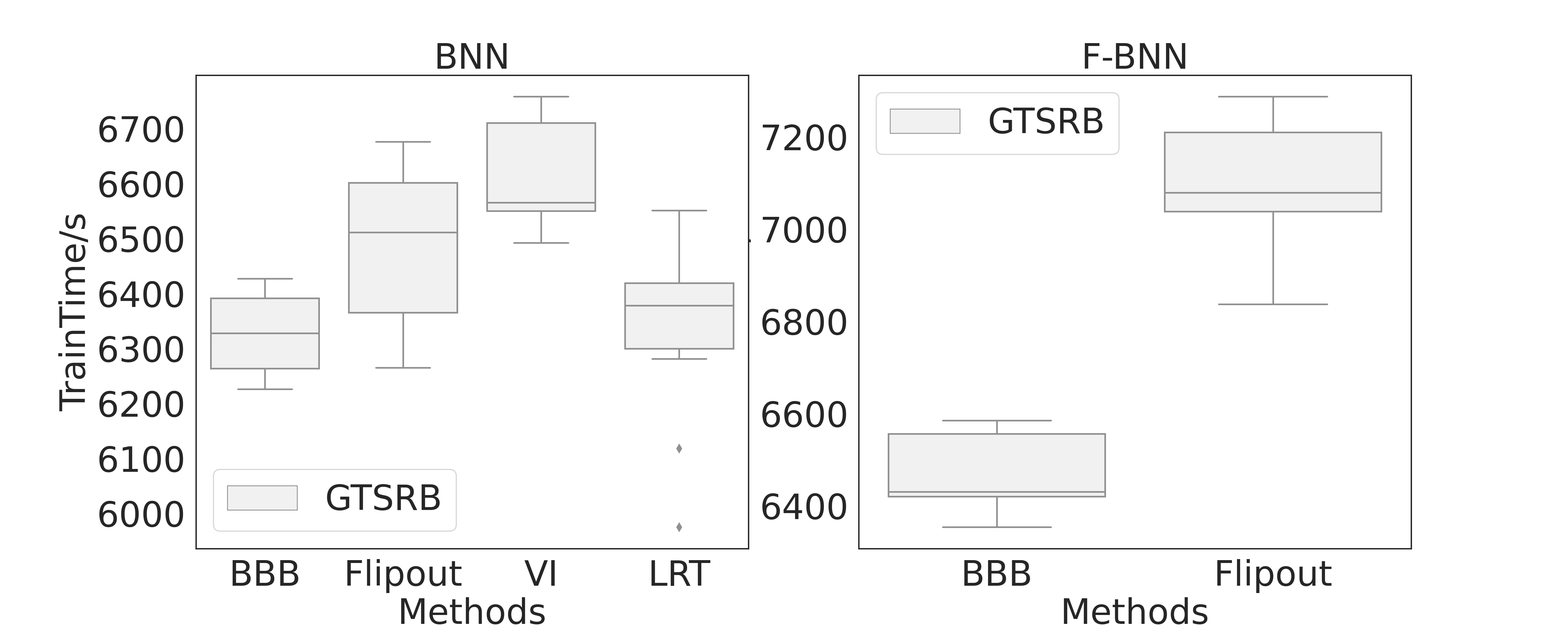}
\caption[Two figures]{Training time comparison on GTSRB with stochastic BNN. \textit{Left: BNN with only classifier as stochastic. Right: F-BNN with stochasticity on each neural network layer.}}
\label{fig: traffic_time_together}
\end{figure}

We also perform training time analysis for BNN and F-BNN with various settings in Figure.~\ref{fig: traffic_time_together}. The model architecture, layer setups, and training procedure for different methods are kept identical to address a fair comparison. The minimum median time used for training BNN is the Bayes By Backprop method, with smaller interquartile range. This also holds true for F-BNN training where only BBB and Flipout are used. We observe that BNN requires less computer training time. The same pattern is discovered for both datasets.

\section{Discussions}

For adversarial attacks, the PGD achieves best attack performance towards our CNN baseline models with reasonable perturbations. However, BNN and F-BNN also exhibit significant robustness gain under PGD attacks. The classification accuracy rises to $80\%$ in F-BNN for GTSRB data. The white-box adversarial samples generated with Eq.~\ref{op2} show bipolarity. This indicates the generation algorithm needs specific parameter tuning. This is beyond the scope of this work. Bayesian NN also proves to be robust under these cases. The peer comparison between stochastic setting shows that BNN and F-BNN are analogous, in the sense of robustness against white-box adversarial attacks. The results shows inconsistency among different types of black-box attacks, especially adversarial samples based on Eq.~\ref{op2}. For instance, with $\varepsilon=0.10$, SPSA shows better attack performance on GTSRB but MIM has the best attack performance on PlanesNet. SPSA fails every model for GTSRB, still due to the large perturbations to original inputs. Bayesian NN shows better performance against black-box attacks. In some cases, F-BNN has slightly better performance compare to BNN but others not. This peer comparison shows similar results with white-box attacks.

Overall, Bayesian NN shows remarkably robustness against all types of adversarial attacks except SPSA on GTSRB. This is due to the large, human-visible perturbations generated from SPSA. Larger adversarial perturbations cased by larger $\varepsilon$ value makes the model perform worse, unless the model has already failed with small perturbations. BNN achieves comparable performance to F-BNN in the sense of classification accuracy, for both white-box and black-box attacks.


\section{Conclusion \label{sec: conclusion}}

We highlight several discoveries here. Firstly, the Bayesian formulation of Neural Network can remarkably improve the performance of deep learning models, especially when dealing with constrained white-box adversarial attacks. Then, we notice that solely a Bayesian classifier is sufficient to improve model robustness. This decreases the time and space complexity with fewer parameter distributions. It's also insignificant when dealing with augmentation based input perturbations since classical CNN has already shows satisfactory denosing capabilities. Lastly, Bayesian neural network may fail. This happens when dealing images with human-visible modifications. 

In the future, it worth looking at the possible reasons of good performance on image classification task. For instance,  benefits of ensemble methods or the feature of Bayesian statistics. Also, it's interesting to look at the model robustness against attacks that are developed specifically toward BNNs (\eg gradient-free adversarial attacks for BNN \cite{yuan2020gradient}). The Bayesian posterior observed from BNN can act as the safety precursor of ongoing malicious activities toward the deploy machine learning systems. This leads to the detection of adversarial samples in cybersecurity.



\clearpage
{\small

}
\clearpage

\onecolumn
\appendix 
\section{Appendix \label{sec: appendix}}

\subsection{Evaluation Results on PlanesNet Dataset}
We list the classification results of PlanesNet Dataset in Table.~\ref{table3} and Table.~\ref{table4}. The PlanesNet Dataset is open-source online. The objective is to classify the existence of aircrafts in surveillance satellite images. The PlanesNet has two class labels indicate \textit{plane} or \textit{not plane}. PlanesNet is another good example to demonstrate the potential benefits to the engineering applications in a safety-critical domain. Table.~\ref{table3} is the results against input perturbations. Table.~\ref{table4} is the results against adversarial attacks.

\begin{table*}[hbt!]
\center
\caption{Performance Against Input Perturbations. Report Test Accuracy in \%.}
\label{table3}

\scalebox{0.85}{
\begin{tabular}{|p{0.05\linewidth}|p{0.07\linewidth}|p{0.1\linewidth}|p{0.1\linewidth}|p{0.1\linewidth}|p{0.1\linewidth}|p{0.1\linewidth}|p{0.1\linewidth}|p{0.1\linewidth}|}
\hline
\rowcolor{lightgray}\multicolumn{9}{|l|}{\textbf{Dataset II: PlanesNet(Detect Aircraft in Planet Satellite Image Chips)}} \\
\hline
        \multicolumn{2}{|l|}{Methods} & Clean & Gaussian & S\&P & Poisson & RE & RE Colorful & Speckle\\ 
        \hline\hline
        CNN & Baseline & $98.19\pm 0.37$ & $98.21\pm 0.31$ & $86.76\pm1.29$ &$98.22\pm0.38$ & $93.16\pm0.60$& $88.99\pm0.80$& $74.76\pm 0.41$\\
        \hline\hline
        \multirow{ 2}{*}{F-BNN} &Flipout & $97.83\pm 0.12$ &$97.79\pm0.21$ &$91.91\pm0.41$ &$97.86\pm0.15$ &$93.06\pm0.24$ &$89.43\pm 0.75$ & $74.80\pm 0.90$\\
        \cline{2-9}
        & BBB &$96.50\pm 0.69$ &$96.59\pm0.55$ &$87.85\pm1.84$ &$96.59\pm0.70$ &$91.14\pm0.92$ &$88.39\pm0.91$ &$75.33\pm0.91$ \\
        \hline\hline
        \multirow{4}{*}{BNN} &Flipout &$98.55\pm0.28 $ &$98.44\pm0.28 $ &$84.06\pm3.02$ &$98.53\pm0.26 $ &$92.21\pm0.57$ &$89.44\pm0.52 $ &$74.79\pm0.23 $ \\
        \cline{2-9}
        &BBB &$98.73\pm0.23$ &$98.65\pm 0.27$ &$83.84\pm2.96$ &$98.71\pm0.25$ &$91.63\pm0.70 $ &$88.98\pm0.67$ &$75.04\pm 0.32$ \\
        \cline{2-9}
        &LRT &$98.88\pm0.12 $ &$98.87\pm0.11$ &$83.00\pm1.88$ &$98.89\pm0.11$ &$92.30\pm0.55 $ &$89.04\pm0.21 $ &$74.78\pm0.12 $ \\
        \cline{2-9}
        &VI &$96.41\pm2.72$ &$96.23\pm2.82 $ &$82.73\pm3.41 $ &$96.34\pm2.67$ &$89.86\pm2.31$ &$87.01\pm1.33$ &$72.96\pm3.25$ \\
        \hline 
\end{tabular}
}
\end{table*}

\begin{table*}[hbt!]
\center

\caption{Performance Against Adversarial Attacks. Report Test Accuracy in \%.}
\label{table4}

\scalebox{0.8}{
\begin{tabular}{|p{0.06\linewidth}|p{0.07\linewidth}|p{0.1\linewidth}|p{0.1\linewidth}|p{0.1\linewidth}|p{0.1\linewidth}|p{0.1\linewidth}|p{0.1\linewidth}|p{0.1\linewidth}|p{0.1\linewidth}|}
\hline
\rowcolor{lightgray}\multicolumn{10}{|l|}{\textbf{Dataset II: PlanesNet(Detect Aircraft in Planet Satellite Image Chips)}} \\
\hline
        \multicolumn{2}{|l|}{$L_{\infty}(\epsilon=.10$)}  & \multicolumn{5}{l}{White-box Attacks} & \multicolumn{3}{|l|}{Black-box Attacks}\\\hline
        \multicolumn{2}{|l|}{Methods/Distance} & PGD/$0.068$ & FGSM/$0.086$ & BIM/$0.061$ & C\&W/$0.005$ & DeepF/$0.349$ & SPSA/$0.212$ & MIM/$0.094$ &Square/$0.080$\\ 
        \hline\hline
        CNN & Baseline & $1.81\pm 0.37$ &$45.44\pm 2.40$ &$13.50\pm2.28 $ &$68.21\pm2.13$ & $43.77\pm4.83$ & $49.65\pm0.91$ &$1.83\pm0.36$ & $15.46\pm1.70$\\
        \hline\hline
        \multirow{2}{*}{F-BNN} &Flipout & $24.51\pm3.70 $ &$57.69\pm2.37 $ &$36.78\pm4.35 $ &$91.13\pm0.41$ & $54.90\pm2.84$ & $60.88\pm0.79$ &$23.14\pm2.71$ & $75.36\pm6.81$\\
        \cline{2-10}
        & BBB & $28.43\pm 5.81$ &$58.51\pm2.65 $ &$40.96\pm6.94 $ &$89.69\pm0.82$ & $48.54\pm8.92$ & $66.54\pm1.06$ &$26.28\pm5.04$ & $76.20\pm7.44$\\
        \hline\hline
        \multirow{4}{*}{BNN} &Flipout & $23.54\pm 2.67$ &$62.77\pm1.30 $ &$40.31\pm 3.33$ &$91.33\pm0.76$ & $68.30\pm1.33$ & $65.58\pm2.68$ &$24.07\pm2.25$ & $75.89\pm2.65$\\
        \cline{2-10}
        &BBB & $22.58\pm4.24 $ &$59.74\pm4.23 $ &$38.69\pm8.71 $ &$91.59\pm1.35$ & $61.85\pm5.06$ & $62.94\pm2.94$ &$22.72\pm4.20$ & $74.98\pm4.11$\\
        \cline{2-10}
        &LRT & $25.22\pm1.80 $ &$62.59\pm1.09 $ &$42.23\pm3.93$ &$91.47\pm1.42$ & $69.33\pm2.28$ & $65.57\pm2.16$ &$25.83\pm2.42$ & $77.79\pm1.59$\\
        \cline{2-10}
        &VI & $20.72\pm5.56 $ &$55.79\pm3.38 $ &$33.92\pm 6.17$ &$91.19\pm1.70$ & $65.89\pm4.76$ & $60.30\pm2.49$ &$19.91\pm5.57$ & $72.56\pm4.02$\\
        \hline 
\end{tabular}
}

\scalebox{0.8}{
\begin{tabular}{|p{0.06\linewidth}|p{0.07\linewidth}|p{0.1\linewidth}|p{0.1\linewidth}|p{0.1\linewidth}|p{0.1\linewidth}|p{0.1\linewidth}|p{0.1\linewidth}|p{0.1\linewidth}|p{0.1\linewidth}|}
\hline
\rowcolor{lightgray}\multicolumn{10}{|l|}{\textbf{Dataset II: PlanesNet(Detect Aircraft in Planet Satellite Image Chips)}} \\
\hline
        \multicolumn{2}{|l|}{$L_{\infty}(\epsilon=.15$)}  & \multicolumn{5}{l}{White-box Attacks} & \multicolumn{3}{|l|}{Black-box Attacks}\\\hline
        \multicolumn{2}{|l|}{Methods/Distance} & PGD/$0.085$ & FGSM/$0.127$ & BIM/$0.084$ & C\&W/$0.007$ & DeepF/$0.352$ & SPSA/$0.217$ & MIM/$0.139$ &Square/$0.113$\\ 
        \hline\hline
        CNN & Baseline & $1.81\pm 0.37$ &$50.96\pm 2.07$ &$12.79\pm2.37 $ &$63.92\pm1.99$ & $45.14\pm4.89$ & $49.03\pm0.67$ &$1.81\pm0.36$ &$12.53\pm2.90$ \\
        \hline\hline
        \multirow{2}{*}{F-BNN} &Flipout & $16.71\pm2.83 $& $56.17\pm1.66$& $20.74\pm2.99$& $89.33\pm 0.86$&$52.86\pm4.97$ & $61.99\pm0.60$ &$17.28\pm3.62$ &$64.91\pm7.99$ \\
        \cline{2-10}
        & BBB &$21.33\pm 5.90$ &$57.48\pm0.57 $ & $23.98\pm3.64$&$87.85\pm1.27$ &$52.00\pm3.38$& $68.57\pm1.92$ &$19.77\pm4.45$ &$68.76\pm7.46$  \\
        \hline\hline
        \multirow{4}{*}{BNN} &Flipout & $19.33\pm4.21 $& $63.91\pm4.22 $& $25.28\pm5.88 $&$90.16\pm0.56 $ &$67.64\pm3.04 $ &$67.35\pm2.32$ & $22.76\pm5.80$ &$72.38\pm1.57$ \\
        \cline{2-10}
        &BBB &$16.85\pm3.95$ &$63.92\pm3.16$ &$24.04\pm3.61$ & $90.32\pm0.84$&$65.43\pm 2.05$& $67.89\pm2.97$ &$19.54\pm4.39$ &$72.58\pm2.80$  \\
        \cline{2-10}
        &LRT & $19.01\pm4.72$&$63.54\pm2.56$ &$24.27\pm4.83$ &$90.85\pm0.34$ &$69.08\pm 1.94$& $67.09\pm1.24$ &$21.52\pm5.07$ &$70.67\pm2.45$  \\
        \cline{2-10}
        &VI &$24.82\pm 7.63$ &$65.39\pm4.98$ &$32.04\pm 9.46$ &$88.21\pm3.27 $ & $68.59\pm 3.37$& $67.99\pm4.05$ &$30.16\pm10.1$ &$72.96\pm1.64$ \\
        \hline 
\end{tabular}
}

\end{table*}

\subsection{Visualization of the Data}
\begin{figure*}[t!]
    \centering
    \begin{subfigure}[t]{0.45\textwidth}
      \includegraphics[width=1\linewidth]{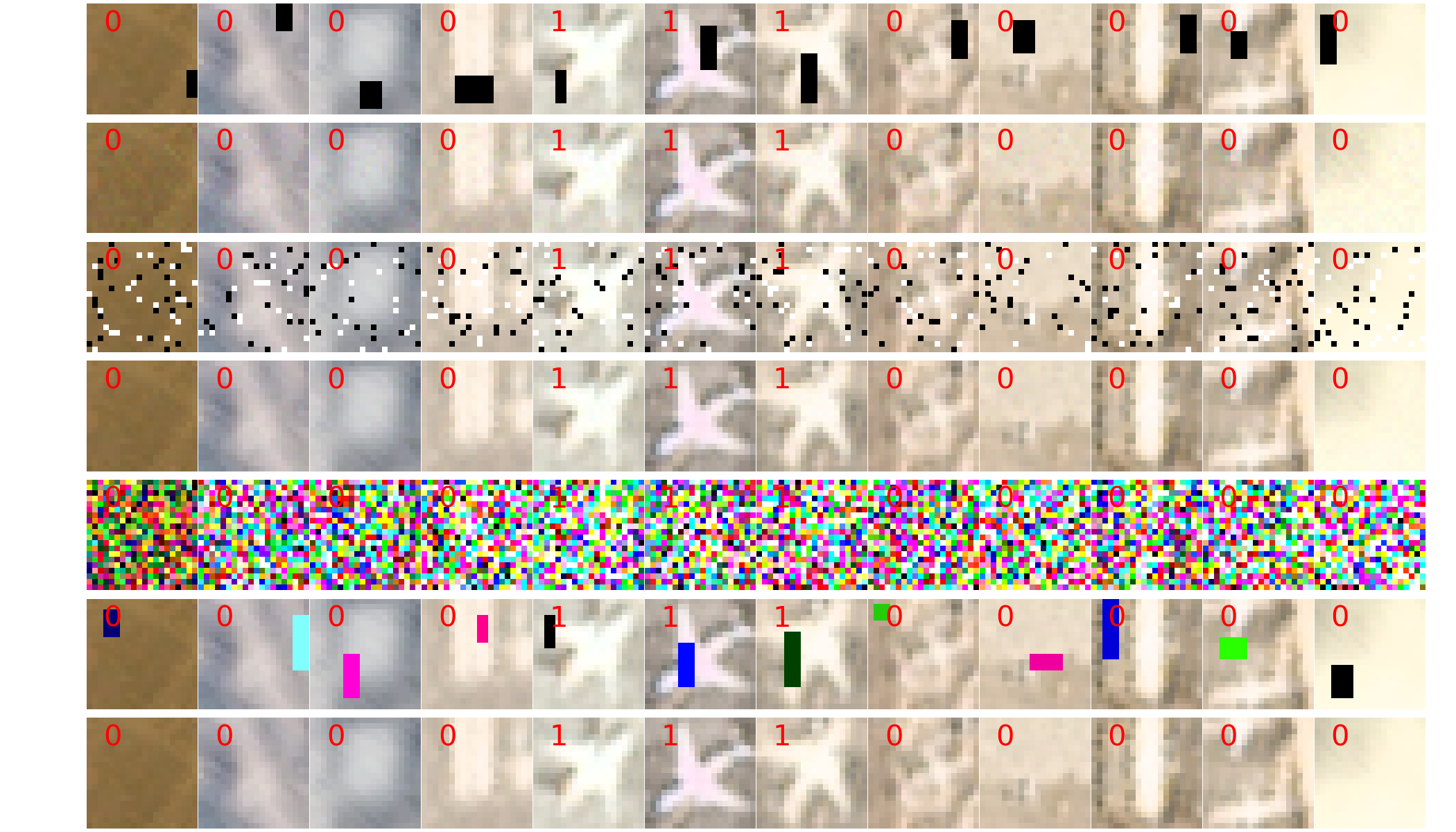}
      \caption{PlanesNet with Input Perturbations}
    \end{subfigure}
    ~ 
    \begin{subfigure}[t]{0.45\textwidth}
      \includegraphics[width=1\linewidth]{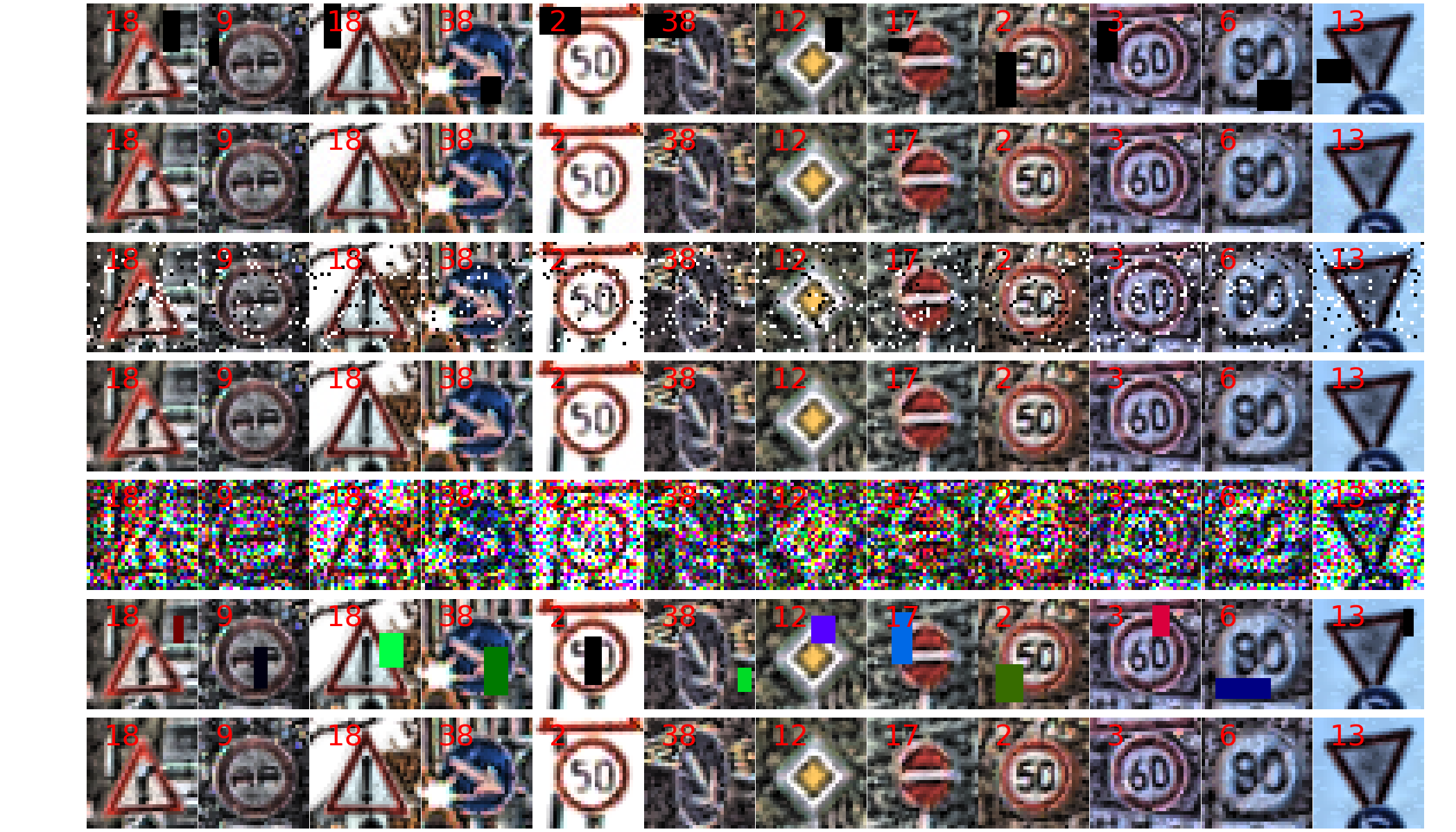}
      \caption{GTSRB with Input Perturbations}
    \end{subfigure}
    
\caption[Two figures]{Visualization of applying input perturbations to PlanesNet and GTSRB. From top to bottom, the methods are: RE, Gaussian, S\&P, Possion, Speckle, RE Colorful, Clean. The ground-truth for class labels are indicated at the top-left corner of each data plot. (a) PlanesNet.  (b) GTSRB}
\label{fig: ips}

\end{figure*}

\begin{figure*}[t!]
    \centering
    \begin{subfigure}[t]{0.45\textwidth}
      \includegraphics[width=1\linewidth]{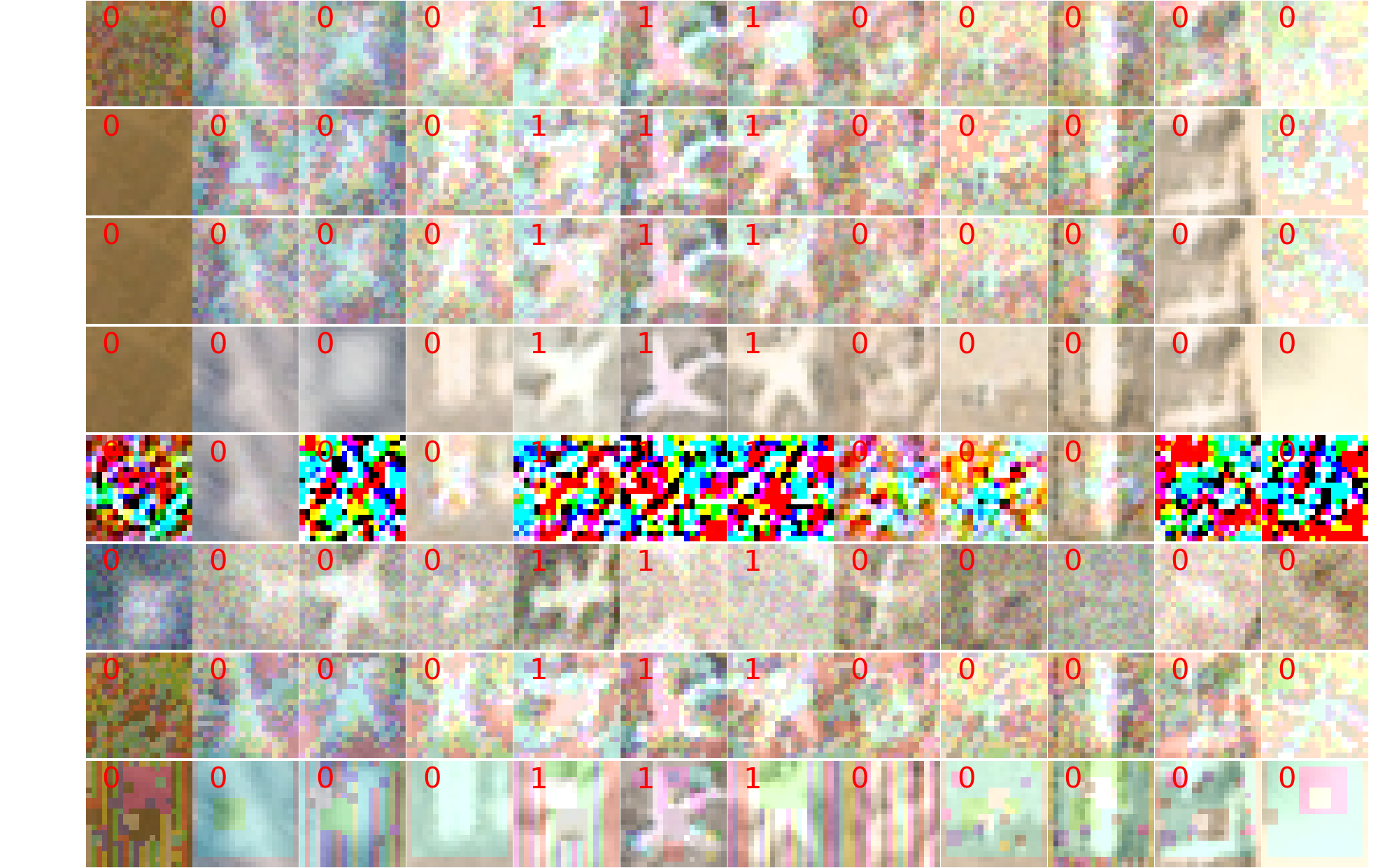}
      \caption{Adversarial Samples for PlanesNet: $\varepsilon=0.10$}
    \end{subfigure}
    ~ 
    \begin{subfigure}[t]{0.45\textwidth}
      \includegraphics[width=1\linewidth]{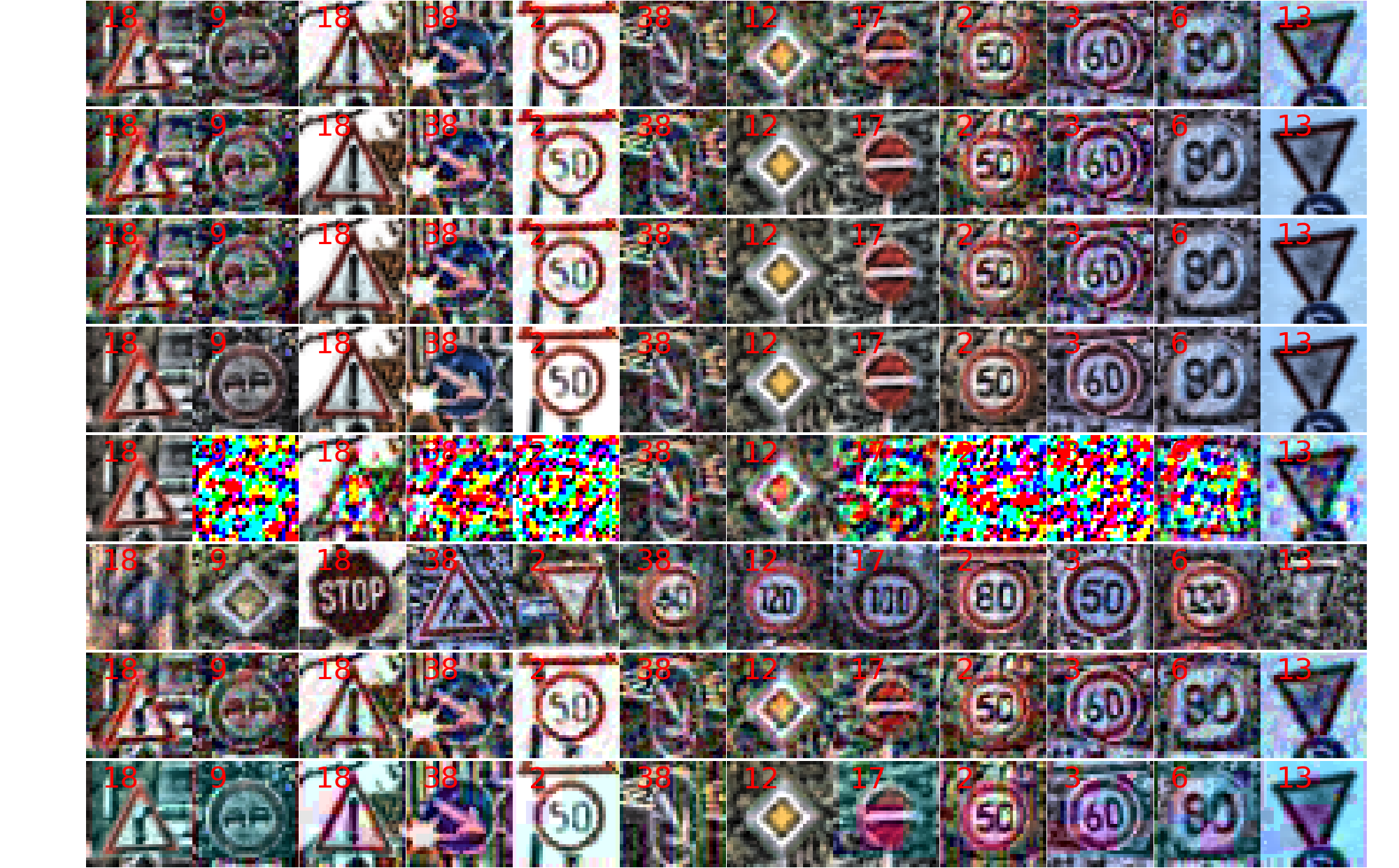}
      \caption{Adversarial Samples for GTSRB: $\varepsilon=0.10$}
    \end{subfigure}
    \\
    \begin{subfigure}[t]{0.45\textwidth}
      \includegraphics[width=1\linewidth]{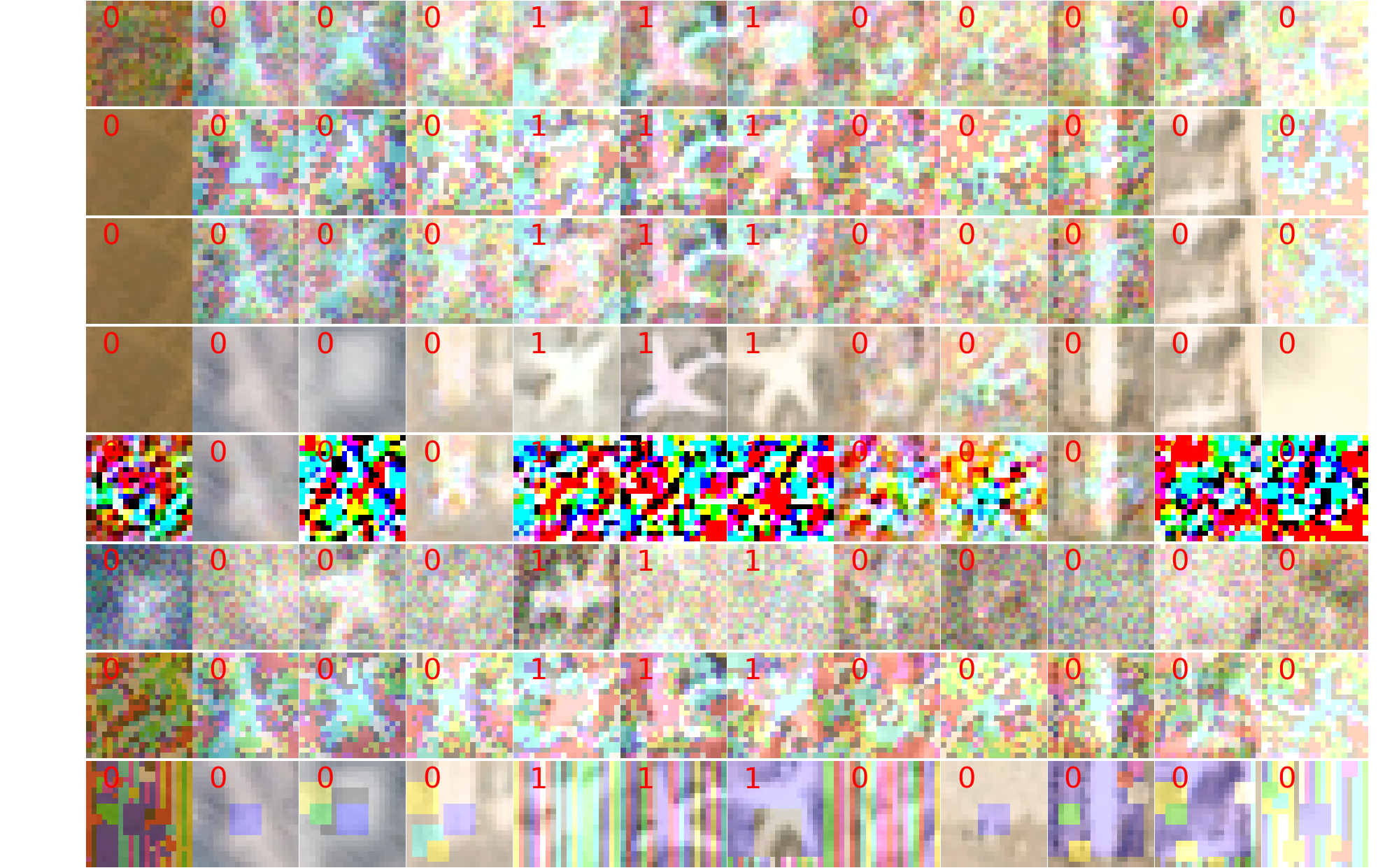}
      \caption{Adversarial Samples for PlanesNet: $\varepsilon=0.15$}
    \end{subfigure}
    ~ 
    \begin{subfigure}[t]{0.45\textwidth}
      \includegraphics[width=1\linewidth]{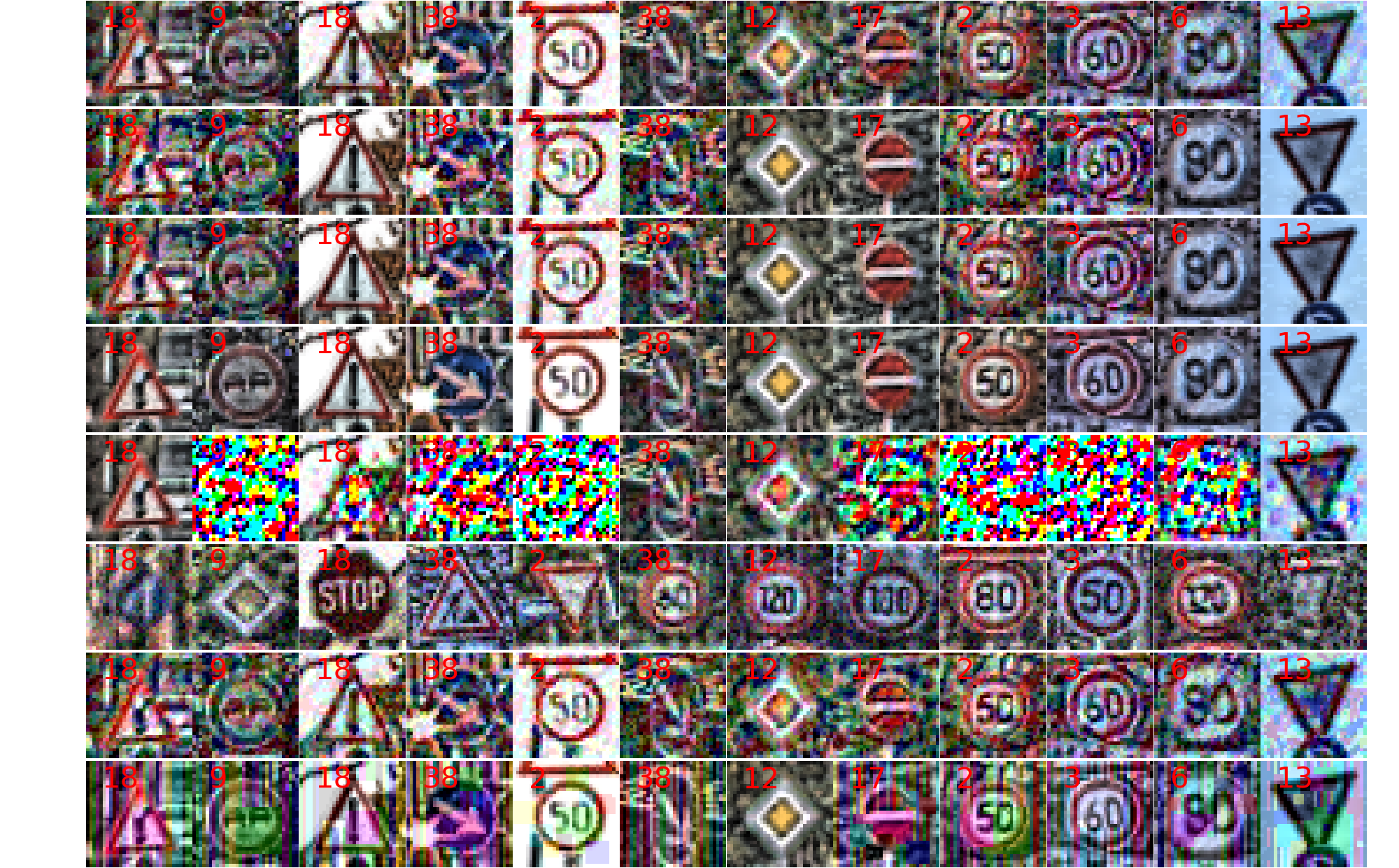}
      \caption{Adversarial Samples for GTSRB: $\varepsilon=0.15$}
    \end{subfigure}
    
\caption[Two figures]{Visualization of applying input perturbations to PlanesNet and GTSRB. From top to bottom, the methods are: PGD, FGSM, BIM, C\&W, DeepF, SPSA, MIM, Square. The ground-truth for class labels are indicated at the top-left corner of each data plot. (a) PlanesNet.  (b) GTSRB}
\label{fig: adv}

\end{figure*}

We visualize the data with input perturbations in Fig.~\ref{fig: ips}. The first 12 samples in the test dataset are chosen with their correct class labels indicated at the top-left corner. The types of input perturbation are Random Erasing, Gaussian, Salt-and-Pepper, Possion, Speckle, Random Erasing Colorful, Clean. Similarly, the adversarial test samples are visualized in Fig.~\ref{fig: adv}, with different perturbation budgets $\varepsilon=0.10$ for (a) \& (b), $\varepsilon=0.15$ for (c) \& (d).


\clearpage
\subsection{Training Time Analysis}
\begin{figure*}[ht!]
\centering
\begin{subfigure}[b]{0.45\textwidth}
  \includegraphics[width=1\linewidth]{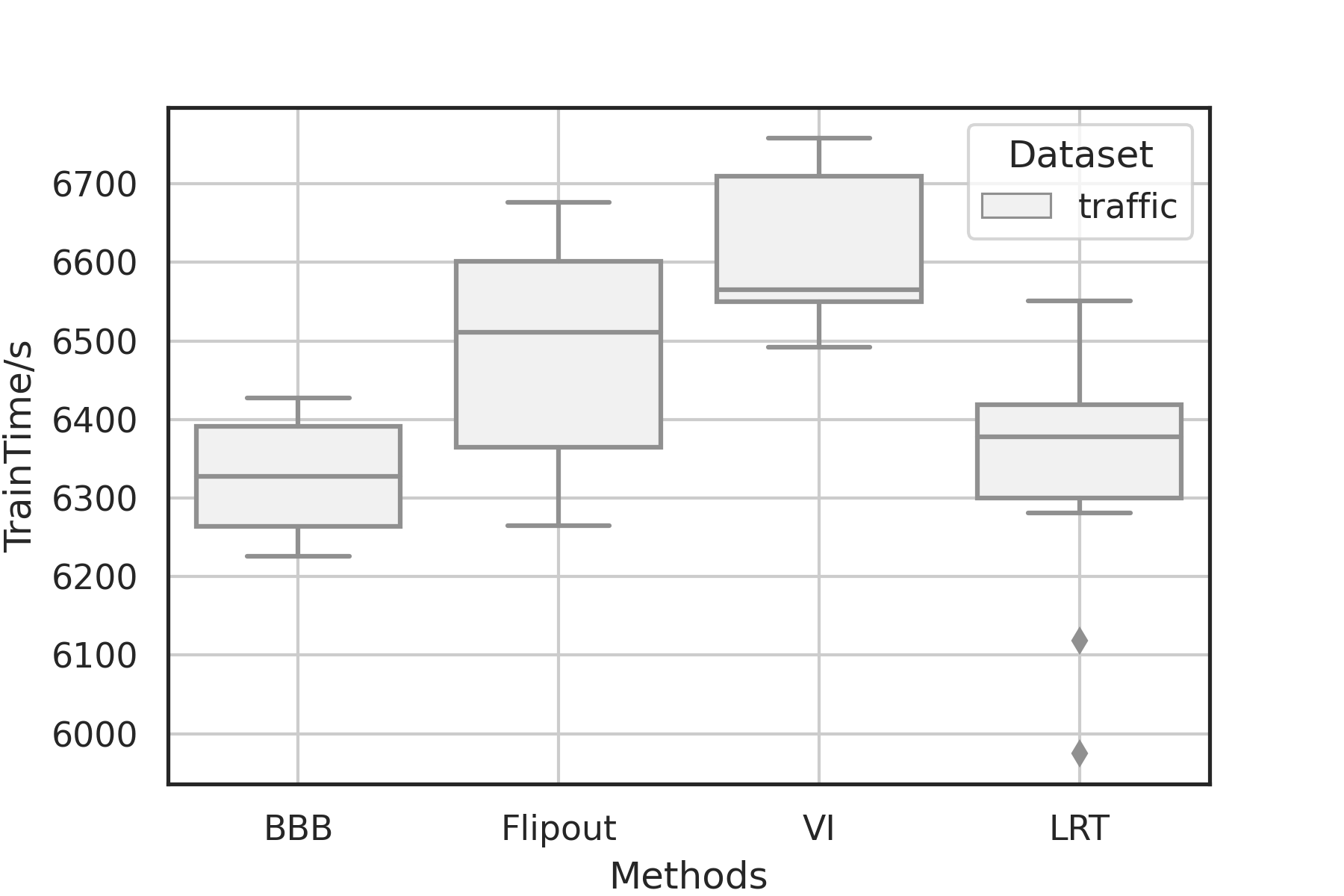}
  \caption{BNN Training time on GTSRB dataset}
  \label{fig:Ng1} 
\end{subfigure}
~
\begin{subfigure}[b]{0.45\textwidth}
  \includegraphics[width=1\linewidth]{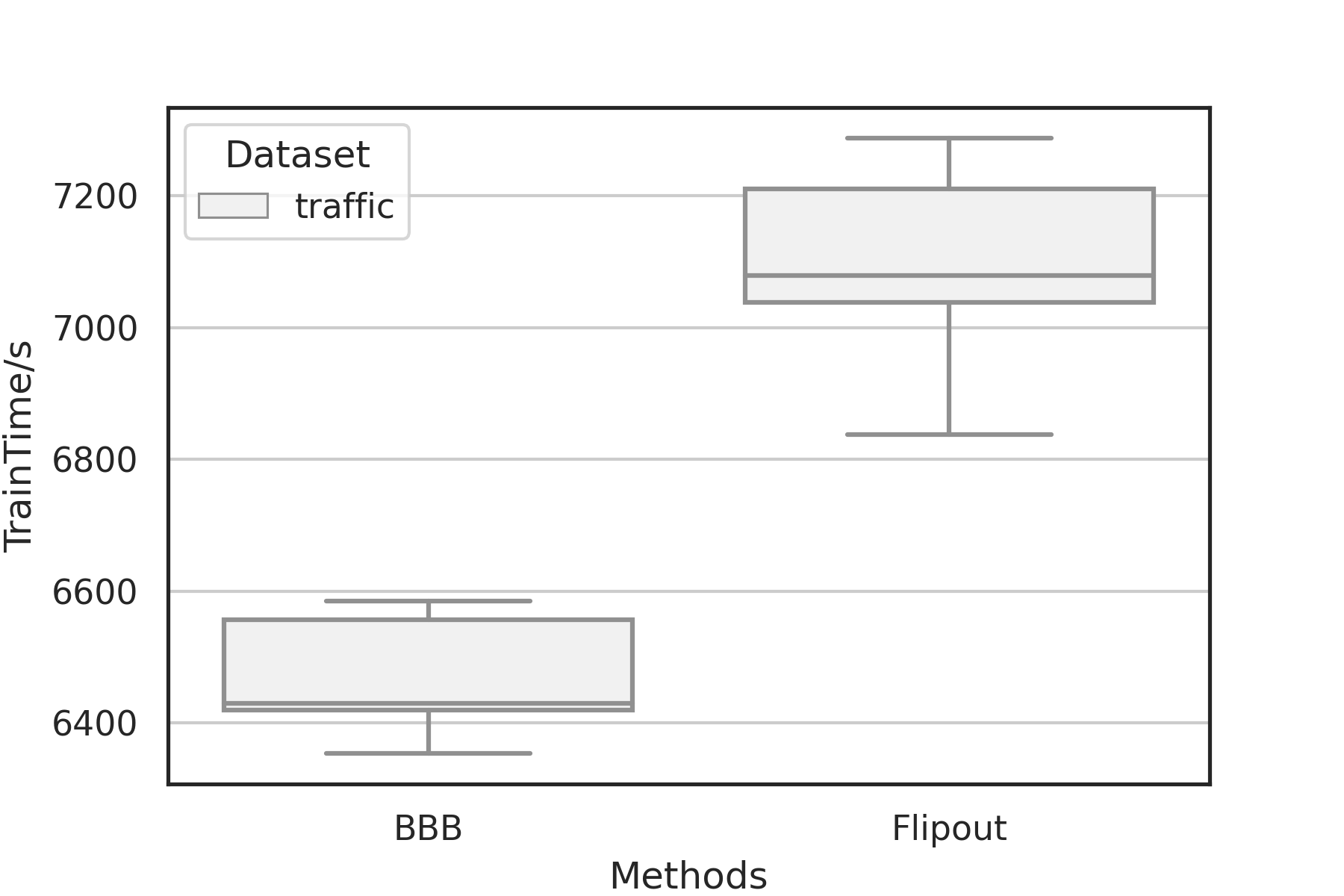}
  \caption{F-BNN Training time on GTSRB dataset}
  \label{fig:Ng2}
\end{subfigure}

\caption[Two figures]{Training time comparison with stochastic BNN. The figures for PlaneNet are reported in Fig.~\ref{fig: planes_time}. (a) BNN with only classifier as stochastic. (b) F-BNN with stochasticity on every neural network layers.}
\label{fig: traffic_time}

\end{figure*}

\begin{figure*}[ht!]
    \centering
    \begin{subfigure}[t]{0.45\textwidth}
      \includegraphics[width=1\linewidth]{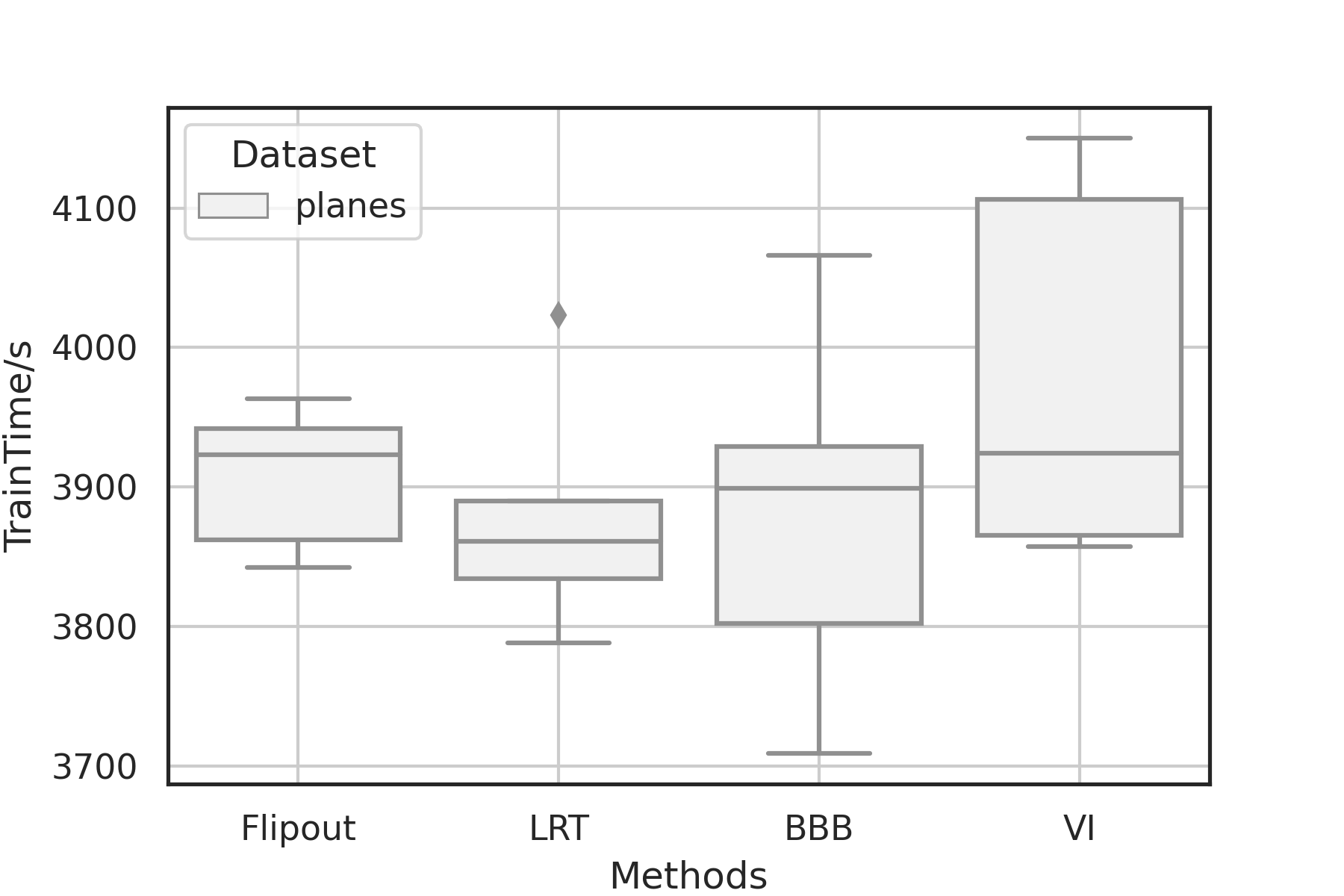}
      \caption{BNN on PlanesNet}
      \label{fig:Ng3} 
    \end{subfigure}%
    ~ 
    \begin{subfigure}[t]{0.45\textwidth}
      \includegraphics[width=1\linewidth]{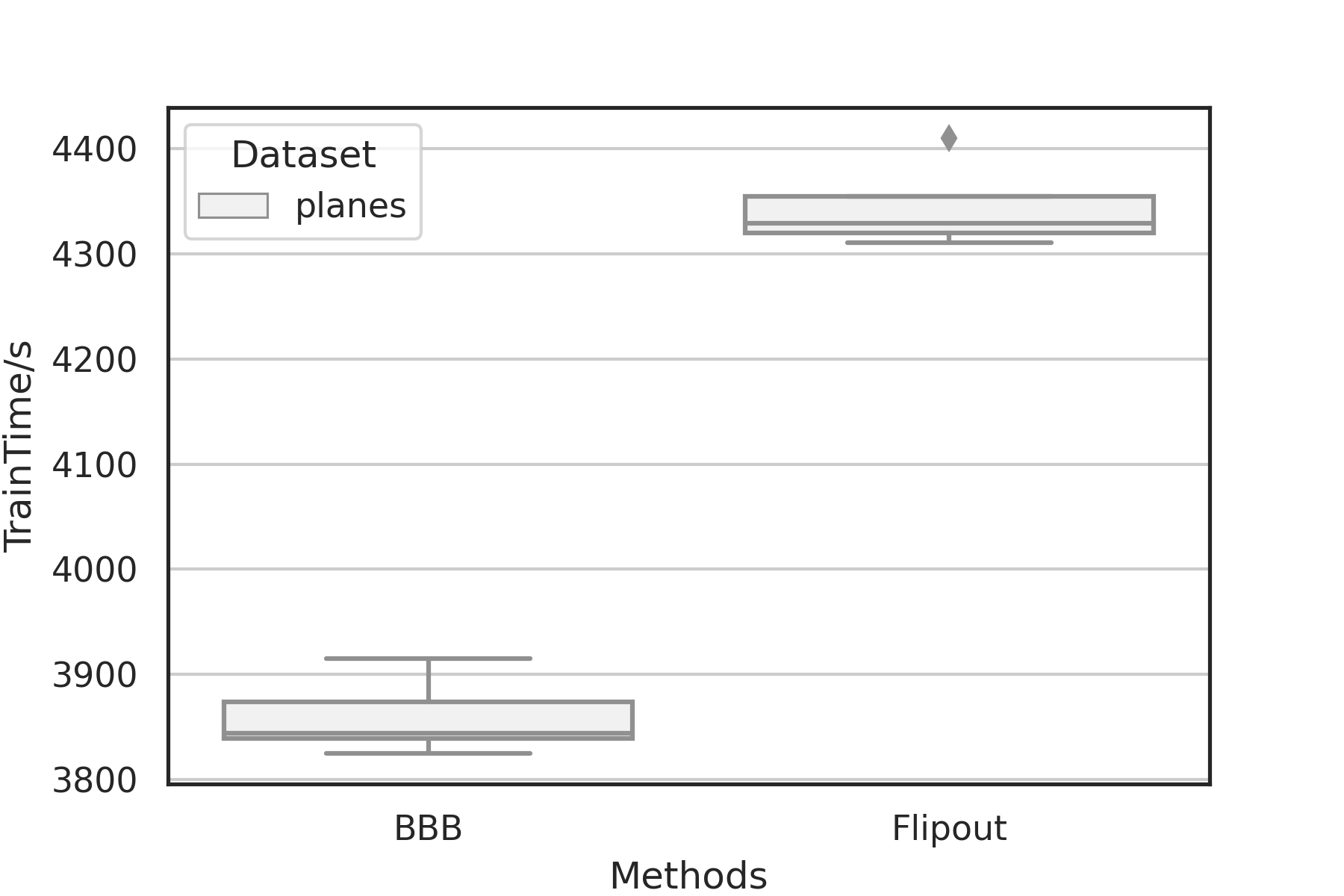}
      \caption{F-BNN on PlanesNet}
      \label{fig:Ng4}
    \end{subfigure}

\caption[Two figures]{Training time comparison with stochastic BNN. (a) BNN with only classifier as stochastic. (b) F-BNN with stochasticity on every neural network layers.}
\label{fig: planes_time}

\end{figure*}

The training time of different methods on both two datasets are compared in Fig.~\ref{fig: traffic_time} and Fig.~\ref{fig: planes_time}. We can see BNNs require less training time.

\end{document}